%% file: main.tex
\documentclass[letterpaper, 10 pt, conference]{ieeeconf}

\IEEEoverridecommandlockouts
\usepackage{amsmath,amssymb,amsfonts}

\usepackage{algorithmic}
\usepackage{graphicx}
\usepackage{textcomp}
\usepackage{xcolor}
\usepackage{newclude}
\usepackage{bm}
\usepackage{caption}
\usepackage[hidelinks, colorlinks]{hyperref}
\usepackage[nameinlink]{cleveref} 
\hypersetup{colorlinks = true,urlcolor=magenta,linkcolor=blue, citecolor=green}
\begin{document}

\title{\LARGE \bf Feature-Realistic Neural Fusion for Real-Time, Open Set Scene Understanding
}

\author{Kirill Mazur, Edgar Sucar and Andrew J. Davison%
\thanks{Dyson Robotics Lab, Imperial College London, UK. Research presented in this paper has been supported by Dyson Technology Ltd.}%
\thanks{}%
}

\maketitle
\thispagestyle{empty}
\pagestyle{empty}

\begin{abstract}
General scene understanding for robotics requires 
flexible semantic representation, so that novel objects and structures which may not have been known at training time can be identified, segmented and grouped.
We present an algorithm which fuses general learned features from a standard pre-trained network into a highly efficient 3D geometric neural field representation during real-time SLAM. The fused 3D feature maps inherit the coherence of the neural field's geometry representation. This means that tiny amounts of human labelling interacting at runtime enable objects or even parts of objects to be robustly and accurately segmented in an open set manner.  Project page: \url{https://makezur.github.io/FeatureRealisticFusion/}
\end{abstract}

\include*{intro}
\include*{method}

\include*{experiments}
\include*{conclusion}

{\small
\bibliographystyle{IEEEtran}
\bibliography{robotvision}
}

\end{document}

%% file: intro.tex
\section{Introduction}

Robots which aim towards general, long-term capabilities in complex environments such as homes must use vision and other sensors to build scene representations which are both geometric and semantic. Ideally these representations should be general purpose, enabling many types of task reasoning, while also efficient to build, update and store.

Semantic segmentation outputs from powerful single-frame neural networks can be fused into dense 3D scene reconstructions to create semantic maps. Systems such as SemanticFusion~\cite{McCormac:etal:ICRA2017} have shown that this can be achieved in real-time to be useful for robotics. However, such systems only make maps of the semantic classes pre-defined in training datasets, which limits how broadly they can be used. Further, their performance in applications is often disappointing as soon as real-world conditions vary too much from their training data.

In this paper we demonstrate the advantages of an alternative real-time fusion method using general learned features, which tend to have semantic properties but remain general purpose when fused into 3D. They can then be grouped with scene-specific semantic meaning in an open-set manner at runtime via tiny amounts of labelling such as a human teaching interaction.
Semantic regions, objects or even object parts can be 
persistently segmented in the 3D map.

In our method, input 2D RGB frames are processed by networks pre-trained on the largest image datasets available, such as ImageNet~\cite{Deng:etal:CVPR2009},
to produce pixel-aligned banks of features, at the same or often lower resolution than the input frames.
We employ either a classification CNN~\cite{Tan:etal:ICML2019} or a Transformer trained in a self-supervised manner~\cite{Caron:etal:ICCV2021}. 
We deliberately use these off-the-shelf pre-trained networks to make the strong point that any sufficiently descriptive learned features are suitable for our approach.

\begin{figure}[t!]
    \centering
    \includegraphics[width=\linewidth, 
                     ]{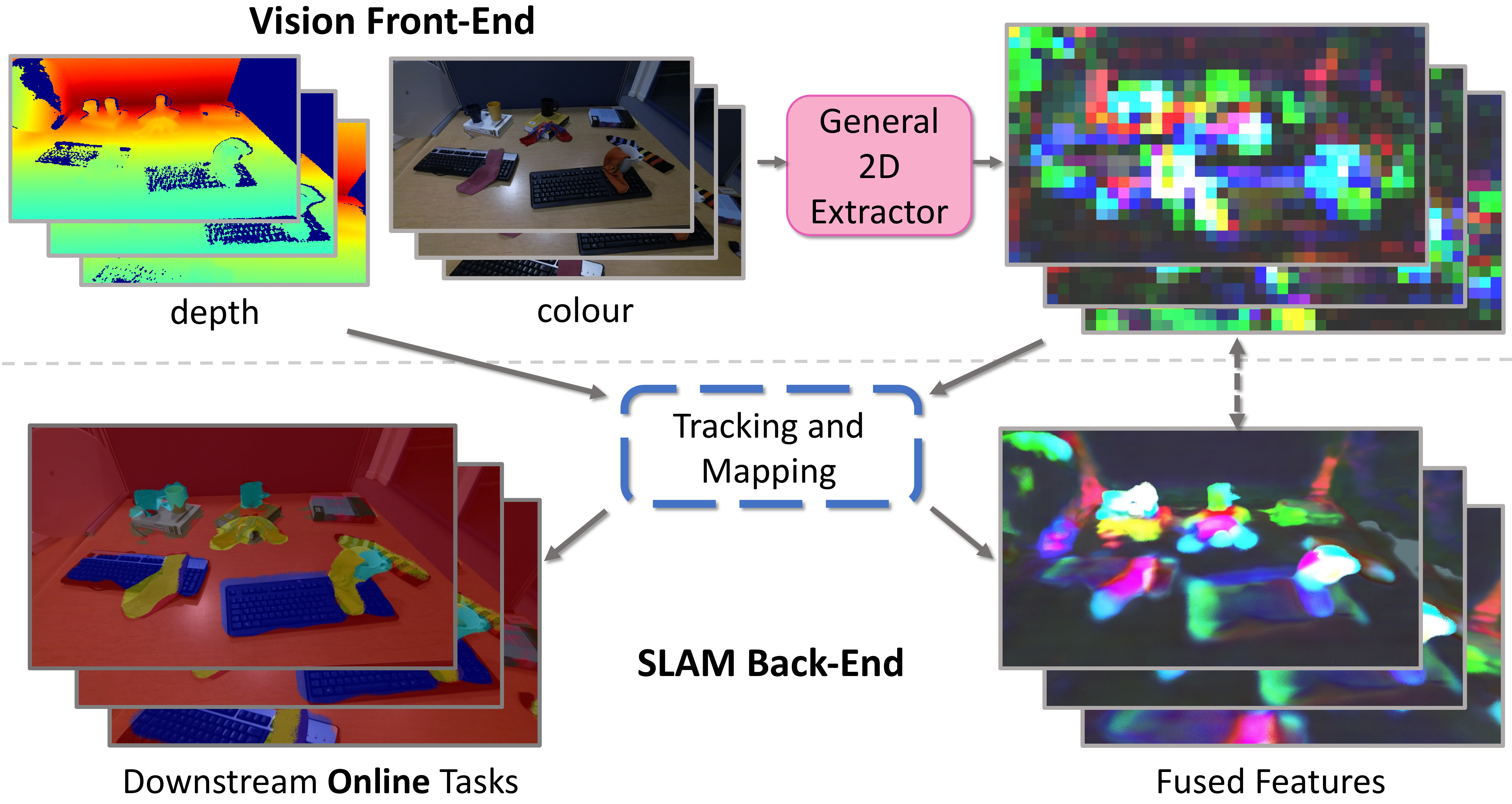}
    \caption{\textbf{Method Overview.} We fuse general pre-trained features into a coherent 3D neural field SLAM model in real-time. The fused feature maps enable highly efficient open set scene labelling during live operation.}
    \label{fig:teaser}
\end{figure}

Rather than fusing features via essentially painting feature distributions onto an explicit 3D geometric reconstruction as is done with semantic classes in~\cite{McCormac:etal:ICRA2017}, here we represent geometry and feature maps jointly via a neural field.
 Neural fields have been recently shown to enable joint representation of geometry and semantics within a single network, such as in the off-line SemanticNeRF system \cite{Zhi:etal:ICCV2021}. The great advantage of this is that the semantic representation 
 inherits the coherence of shape and colour reconstruction, and this means that semantic regions 
 can accurately fit the shapes of objects
 even with very sparse annotation.

We base our new real-time neural feature fusion system on iMAP~\cite{Sucar:etal:ICCV2021}, a neural field SLAM system which uses RGB-D input to efficiently map scenes up to room scale.
We augment iMAP with a new latent volumetric rendering technique, which enables fusion of very high dimensional feature maps with little computational or memory overhead.

We call our scene  representation ``feature-realistic'' as a counterpoint to the ``photo-realistic'' scene models which are the aim of many neural field approaches such as NeRF~\cite{Mildenhall:etal:ECCV2020}. We believe that robotics usually does not need scene representations which precisely model the light and colours in a scene, and that it is more valuable and efficient to store abstract feature maps which relate much more closely to semantic properties.

We demonstrate the scene understanding properties of our system via an open-set semantic segmentation task with sparse user interaction, which represents the way a human might interact with a robot to efficiently teach it about a scene's properties and objects.
The user uses a few pointing clicks to give labels to pixels, and the system then predicts these label properties for the whole scene. We show that compelling dense 3D scene semantic mapping is possible with incredible sparse teaching input at runtime, even for object categories which were never present in training datasets. Usually the user only needs to place one click on an object of a certain type for all instances of that class to be densely segmented from their surroundings.
We evaluate the system on a new custom open-set video segmentation dataset.

To summarise, our contributions are as follows:
    \begin{itemize}
        \item The first neural field feature fusion system operating in \textit{real-time};
        \item A system that operates \textit{incrementally} and successfully handles exploration of previously unobserved scene regions; 
        \item A \textit{latent volumetric rendering} technique which allows fusion up to \textbf{1536}-dimensional feature-maps with negligible performance overhead compared to iMAP and a scene representation of only 3 MB of parameters;
        \item Dynamic open set semantic segmentation application of the presented method;
    \end{itemize}

\section{Related work}

SemanticFusion~\cite{McCormac:etal:ICRA2017}, an extension of ElasticFusion~\cite{Whelan:etal:RSS2015}, introduced a mechanism to incrementally fuse 2D semantic label predictions from a CNN into a three-dimensional environment map. Among other similar systems, the panoptic fusion approach of~\cite{Narita:etal:IROS2019} made an advance by explicitly representing object instances alongside semantic region classes. The latest systems in this vein wield neural fields as an underlying 3D representation. The advantageous properties of the coherence of neural fusion were first shown by Semantic NeRF~\cite{Zhi:etal:ICCV2021}, with variations aimed towards multi-scene generalisation and panoptic fusion demonstrated in~\cite{fu:etal:3DV2022, Kundu:etal:CVPR2022}.

The aforementioned methods suffer from a training/runtime domain gap and the inherently closed-set nature of a fixed semantic label set. The domain is fixed by the dataset and the closed target label set employed during the semantic segmentation model pre-training.  

Our method relates to two recently released approaches, Distilled Feature Fields (DFF)~\cite{Kobayashi:etal:ARXIV2022} and Neural Feature Fusion Fields (N3F)~\cite{Tschernezki:etal:3DV2022}, which also add a feature output branch to a neural field network and supervise the renders with the outputs of a pre-trained feature extractor. 

Unlike our work, N3F and DFF supervise neural fields with up to $64$- and $384$-dimensional feature maps respectively, which is $24\times$ and $5\times$ times smaller than our proposed method. Both DFF and N3F operate in an off-line protocol similar to NeRF and require approximately a day to converge on a single scene, whereas our system operates at \textit{interactive frame rates} making it useful for robotics. Additionally, N3F heavily leverages offline assumptions on an input sequence: all frames have to be known prior to training, due to a pre-processing step which executes dimensionality reduction jointly on all input feature maps. In our online execution paradigm these assumptions would be fundamentally violated and the input distribution might change drastically in a few seconds (e.g. entering a new room).

Both N3F and DFF mainly consider object retrieval and 3D object segmentation mask extraction scenarios. In contrast, we focus on extracting \textit{all} object instances of varying appearance and geometry, given a semantic class. While DFF also considers the semantic segmentation scenario, it fuses the penultimate activations of a pre-trained semantic segmentation model. This method is therefore essentially equivalent to a SemanticNeRF-style approach with the same benefits and pitfalls, such as the domain gap.

Our method achieves real-time performance by using a core neural field SLAM approach based on iMAP~\cite{Sucar:etal:ICCV2021}, with a small MLP network, RGB-D input and guided keyframe and pixel sampling for efficiency. This type of efficient network is well suited to semantic and label fusion. Recent work iLabel~\cite{Zhi:etal:arxiv2021}, also based on iMAP,
showed a type of interactive scene segmentation based on no prior training data. The coherence of the neural field alone was shown to be a basis for segmenting objects from sparse interaction. However, in iLabel there was little evidence that annotation of an object led to grouping with other instances of the same class. In our work we specifically show that this becomes possible due to fusion of general features from an off-the-shelf pre-trained network.

Our method also closely relates to SemanticPaint~\cite{Valentin:etal:ACMTOG2015}, an older online interactive labelling system. 
SemanticPaint, like our system, operates by propagating user-given labels to novel object instances. However, propagation is severely limited to objects which are almost identical apart from  colour. The core of the SemanticPaint  is a random forest classifier with hand-crafted features and refinement with a Conditional Random Field. This machinery cannot compete in pattern recognition abilities with the modern deep learning methods for computer vision our approach builds on. Our system benefits both the best properties of neural fields which encourage coherent segmentation~\cite{Zhi:etal:arxiv2021}, and the power of features from general pre-trained networks.

%% file: method.tex
\section{Method}
\begin{figure}[h]
    \centering
    \includegraphics[width=\linewidth, 
                     ]{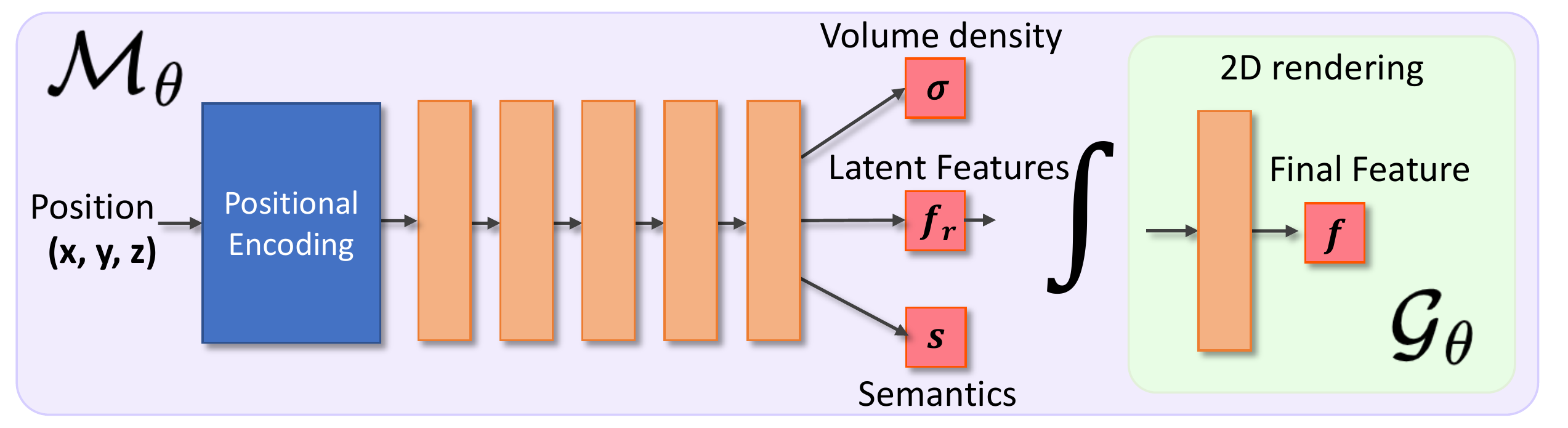}
    \caption{\textbf{Scene Network.} Overview of our Scene Network. Our scene MLP predicts semantics and latent features, which are further refined after the volumetric rendering.}
    \label{fig:method}
\end{figure}

Our system is composed of two principal components: a pre-trained frozen 2D image feature extractor (vision \textit{front-end}) and an iMAP-like SLAM system (SLAM \textit{back-end}). While our method technically allows an image feature extractor of any choice, we focus on ones that are general, i.e not trained for dense prediction tasks.

Our general approach is to approximate via volumetric rendering a set of feature maps $\{\bm{F}_i= \mathcal{F}(\bm{I_i}) \in \mathbb{R}^{k \times H' \times W'} \}_i$ obtained with a feature extractor $\mathcal{F}$ from a set of images $\{ \bm{I_i} \in \mathbb{R}^{3 \times H \times W} \}_i$, similar to~\cite{Kobayashi:etal:ARXIV2022, Tschernezki:etal:3DV2022}. Unlike those methods, we abstain from modelling colour and view-dependant effects to ease the problem for the scene MLP. While colour may seem a more compact representation compared with a high-dimensional feature map, it inherently contains ``nuisance'' variation (due to e.g. illumination changes or camera settings such as auto-exposure), which is usually not relevant to semantic understanding of the scene.  

\subsection{Scene Network}

The architecture of our scene mapping system is largely based on iMAP, whose notation we follow~\cite{Sucar:etal:ICCV2021}. 
Our scene representation network $\mathcal{O}_{\theta} = (\mathcal{M}_{\theta}, \mathcal{G}_{\theta})$ has two components: a NeRF-style multi-layer perceptron $\mathcal{M}_{\theta}$ which serves as a ``scene map'' and represents a three-dimensional neural field; and a single layer perceptron $\mathcal{G}_{\theta}$ which operates in 2D and upsamples volumetrically rendered features to the target dimension $k$ (see \Cref{sec:latent_rendering}).

The coordinate MLP $\mathcal{M}_{\theta}$ with a hidden layer dimension $h=256$ maps a 3D position $\bm{p} = (x, y, z) \in \mathbb{R}^3$ into $\mathcal{M}_\theta(\bm{p}) = (\rho, \bm{f}, \bm{s})$, where $\rho$ stands for volumetric density, $\bm{f}$ for a feature vector, and $\bm{s}$ for semantic logits. Before feeding a point $\bm{p} = (x, y, z)$ into the scene network $\mathcal{M}_{\theta}$ we apply off-axis positional encoding~\cite{barron:etal:CVPR2022} to ensure rich representational capacity for fitting high-frequency feature maps and mitigate axis-aligned artifacts caused by the standard positional encoding used in NeRFs.

We transform densities $\rho_i$ into ray termination probabilities $o_i = 1 - \exp( -\rho_i \delta_i)$ and further into volumetric rendering weights $w_i = o_i \, \Pi_{j=1}^{i - 1} (1 - o_j)$, where $\delta_i$ is the inter-sample distance in the volumetric integration quadrature. We render depth $\hat{D}$, features $\hat{F}_r$, and semantic logits $\hat{S}$ as:

\begin{equation}
    \hat{D}[u,v] = \sum_{i=1}^N w_i d_i, \ \, \hat{F}_{r}[u, v] = \sum_{i=1}^N w_i \bm{f}_i, \ \, \hat{S}[u,v] = \sum_{i=1}^N w_i \bm{s}_i
\end{equation}

We adopt iMAP's key framing strategy: we add a frame into the keyframe set if the current depth relative error is higher then a threshold for more than $65\%$ of pixels. 

The spatial resolution of the feature maps produced by a feature extractor is usually smaller than that of original colour and depth image. To mitigate this we employ a sparse supervision technique for features, similar to the one introduced in SemanticNeRF~\cite{Zhi:etal:ICCV2021}, to let the scene network learn the appropriate feature spatial interpolation. To supervise depth pixels we employ random uniform sampling across the full image resolution for each mapping step.

\subsection{Latent Feature Rendering}
\label{sec:latent_rendering}
NeRFs are well known for their training and inference inefficiency due to the cubic complexity of volumetric rendering. This has led to an extensive line of work endeavoring to alleviate this issue, with recent highlights such as Instant NGP which adds grids and hashing to the MLP representation~\cite{Mueller:SIGGRAPH2022}. 

However, here we choose to stick with a simple MLP as our master scene representation, because of its attractive compression and coherence properties~\cite{Zhi:etal:arxiv2021}

To render a single feature image of spatial dimension $H' \times W'$ a scene network has to be queried $H' \times W' \times N_{\text{bins}}$ times, where $N_{\text{bins}}$ stands for the number of samples per ray. Therefore, given a feature map of  $k$-dimensional features, the memory requirements scales as $k \times H' \times W' \times N_{\text{bins}}$. When the dimensionality of the target features is an order of magnitude higher than the hidden scene MLP dimension ($1536$ and $256$ respectively in our case) the na\"ive approach becomes intractable, especially for a real-time system. 

Our solution is to render a \textit{latent} $h$-dimensional feature vector $\hat{F}_r[u, v]$ followed by a per-point perceptron $\mathcal{G}$ applied \textit{after} the rendering:
\begin{equation}
    \hat{F}[u, v] =  \mathcal{G}_{\theta} \left( \sum_{i=1}^N w_i \bm{f}_i \right)
\end{equation}
This simple approach enables our system to yield up to $k=1536$ dimensional features with a negligible performance and memory overhead.

\subsection{Feature Extractors}

We have observed that models producing highly view-dependent features (such as a standard ViT~\cite{dosovitskiy:etal:ICLR2021}) are unsuitable for our application; view-equivariant effects effectively lead to an underconstrained problem for on-the-fly semantics extraction due to the incremental and online nature of our system. While most ConvNets inherit a shift invariance property from convolution and yield satisfactory feature maps, Transformer-based models tend to be highly equivariant due to the absence of inductive biases. An interesting exception is the Transformer-based DINO~\cite{Caron:etal:ICCV2021} model, pre-trained in an unsupervised setting with a learning objective to produce features invariant to a large set of image transformations. 

In this work we test both convolution-based~\cite{LeCun:etal:1998} and a Transformer-based~\cite{Vaswani:etal:NIPS2017} models to demonstrate that our method is agnostic to the nature of 2D feature front-end and is still capable of fusing these features. The feature extractor network runs inference at the same FPS as  the mapping process (2 Hz, the same as iMAP) to save computation.  

For our ConvNet representative model we choose EfficientNet~\cite{Tan:etal:ICML2019}, a supervised CNN trained on ImageNet~\cite{Deng:etal:CVPR2009}. Each image is passed through the pre-trained network in its original resolution and the output of the final convolutional layer is taken as the target for our system. This process yields a coarse feature map of spatial dimension $22 \times 38$ and each pixel contains a $k=1536$-dimensional feature vector. To mitigate the artifacts caused by padding~\cite{alsallakh:etal:ICLR2021} we ignore a $1 \times 2$-pixel wide frame.

Our Transformer-based model is DINO pre-trained on the ImageNet corpus in a self-supervised manner. We employ the smallest model variant with output feature dimension $k=384$ to achieve real-time performance and fit into the GPU memory. An image is dissected onto patches of size $16 \times 16$ and then fed through the Transformer network.

Our mapping network feature branch is supervised with an $L2$ distance loss in both cases.

%% file: experiments.tex
\section{Experiments}
\begin{figure*}[h!]
  \includegraphics[width=\textwidth,
  trim={0 0cm 7cm 0},clip]{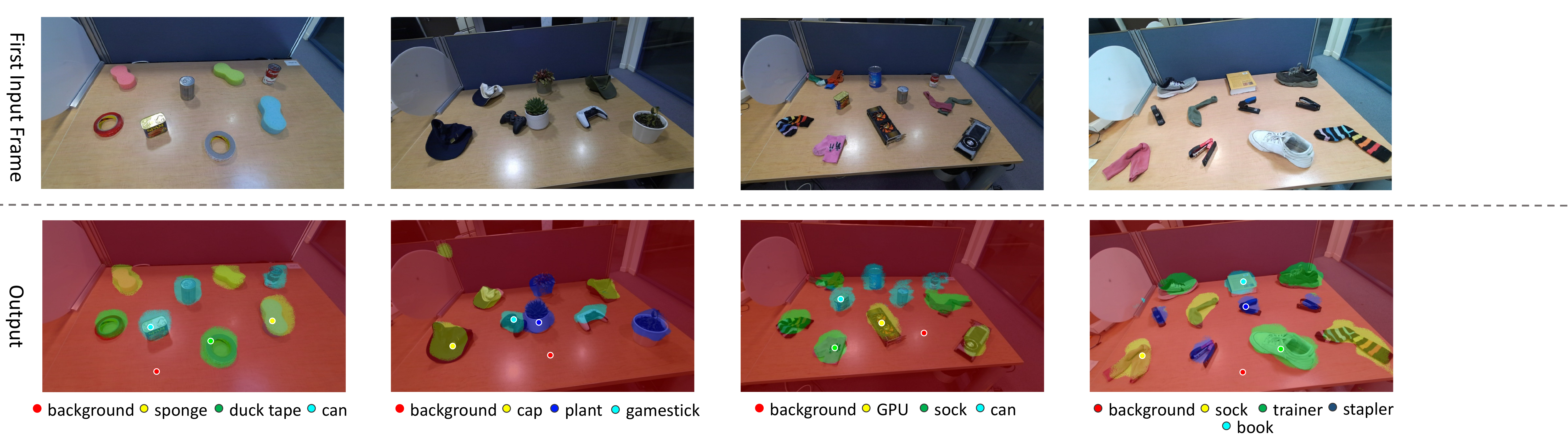}
  \caption{\textbf{Coverage experiments.} Each RGB frame is the first of an RGB-D sequence from which we reconstruct the 3D scene and fuse features in real-time. The coloured dots in the output view are the {\em only} semantic annotations supplied, as clicks on the first frame. Dense semantic predictions are shown, showing high quality semantic segmentation and grouping within instance classes.
  Note that these results use a \textit{very} coarse $22 \times 38$ CNN feature map.}
  \label{fig:coverage}
\end{figure*}

Given a real-time RGB-D video stream $\{(\bf{I}_i, \bf{D}_i)\}$, our system gradually fuses the incoming feature maps $\{\bm{F}_i= \mathcal{F}(\bm{I}_i) \}$ from the front-end and tracks itself in the scene using the SLAM back-end. A user then gradually introduces new classes with ``clicks'' to label \textit{single} pixels in one of the system's keyframes. These labels are used to supervise the semantic head of the scene network $\mathcal{O}_{\theta}$, which results in a 3D semantic segmentation field, similar to SemanticNeRF~\cite{Zhi:etal:ICCV2021}. Inspired by iLabel~\cite{Zhi:etal:arxiv2021} we use interactive labelling as a way to define ``tasks'' for our system on the fly.
This lightweight interaction is representative of how a human might interact with a running robot system to efficiently teach it about named scene properties; or it could represent experimental scene interactions that a robot could carry out for itself.
Note that our system operates in a \textit{one-click-per-class} mode: every click defines a \textit{new} semantic class.  This is unlike iLabel, where several clicks are usually needed to identify large objects or multiple instances of a class.

The system therefore is expected to propagate the semantic label of a click across relevant scene regions, e.g multiple instances of the same object or object parts. Since we stick to a one click-per-class execution protocol, our system is not a labelling system but rather a method to reveal the already present scene part similarities.  

We qualitatively evaluate our system in three experiments: 
\begin{itemize}
    \item \textbf{Coverage:} Grouping objects in rare class scenarios (a sock, a trainer, or a GPU), where traditional models require re-training for a novel class distribution;
    \item \textbf{Specialization}: The ability of our system to specialize from a holistic object category into object part categories, e.g from a mug category into two separate mug handle and body classes;
    \item \textbf{Exploration}: Given a labelled and reconstructed part of the scene, how well labels propagate to previously unreconstructed regions. 
\end{itemize} 

We strongly encourage watching our accompanying \textbf{qualitative demo video}.  
We also quantitatively evaluate the semantic segmentation quality of our method against baselines on our tabletop dataset.

In most experiments we employ the EfficientNet CNN feature front-end unless stated otherwise. We chose the CNN front-end due its ability to provide stronger semantic entanglement compared to its DINO counterpart; see \Cref{sec:cnn_vs_vit}. 

\subsection{Dataset}
Due to the absence of available RGB-D tabletop video datasets which contain repetitive semantic objects, particularly from the unusual classes where the performance of our method is especially notable, we chose to collect our own dataset. It consists of $8$ sequences in total and captured with a handheld Microsoft Azure Kinect camera. The dataset incorporates common household and office objects, such as books, keyboards, trainers, socks, and plants as well rare objects, such as gamesticks and GPUs. 

We randomly sampled ${\sim}5\%$ of the frames per video sequence for five sequences (the sequences from~\Cref{fig:coverage} and \Cref{fig:exploration}) and then densely annotated them with ground-truth labels to obtain the quantitative results in~\Cref{tab:quantitative}.

\subsection{Coverage}
\label{sec:coverage}

\begin{figure}[h]
    \centering
    \includegraphics[width=\linewidth, 
                     ]{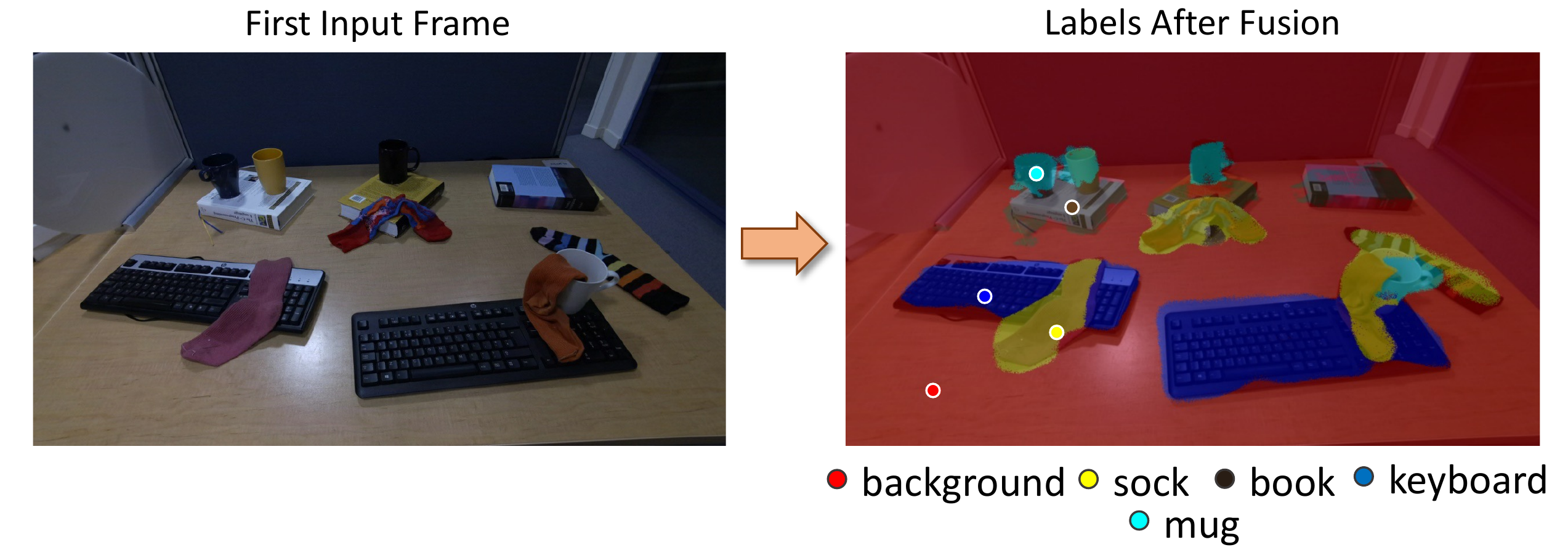}
    \caption{\textbf{Cluttered scene.} Our method successfully handles even cluttered scenes with objects which are in contact and occluding each other.}
    \label{fig:clutter}
\end{figure}

The first set of experiments focuses on evaluating semantic class coverage. In other words, to what extent is our method able to propagate semantic labels from one object  to other instances of the same class. The object semantic categories present in our testing scenes are typically not covered by off-the-shelf models, such as a pre-trained Mask~R-CNN~\cite{He:etal:ICCV2017}. Furthermore, most of these classes are not present as a target class in the ImageNet dataset, on which our EfficientNet CNN was pre-trained. 

Most sequences we use are captured such that the first frame contains the whole scene. We choose this approach to ease the qualitative evaluation, so that all target objects are visible. Our system is also conceptually capable of capturing inherently 3D scenes, which do not fit into a single frame. We additionally cover such cases in \Cref{sec:exploration}. 

In \Cref{fig:coverage} we demonstrate our method's performance after executing it on four tabletop sequences. We observe that our method is capable of propagating semantic class labels across a variety of objects and produces plausible semantic object masks. Note that the method extracts these masks despite fusing very coarse initial CNN feature maps of spatial dimension $22 \times 39$.  

These experiments show that our method particularly shines on unusual object categories, which are typically not present in traditional densely annotated datasets.

\subsection{Specialization}
\label{sec:specialization}
Another potential benefit of our approach not being restricted to a fixed class set is the ability to split classes further down the natural semantic hierarchy. To test out this property we captured scenes with several instances of semantically composite objects, mugs and headphones. We also equip the scenes with an additional object of unrelated class (book) to ensure that the observed label propagation is not incidental.

\begin{figure}[h]
    \centering
    \includegraphics[width=\linewidth, 
                     trim={0 1.5cm 1cm 0},clip]{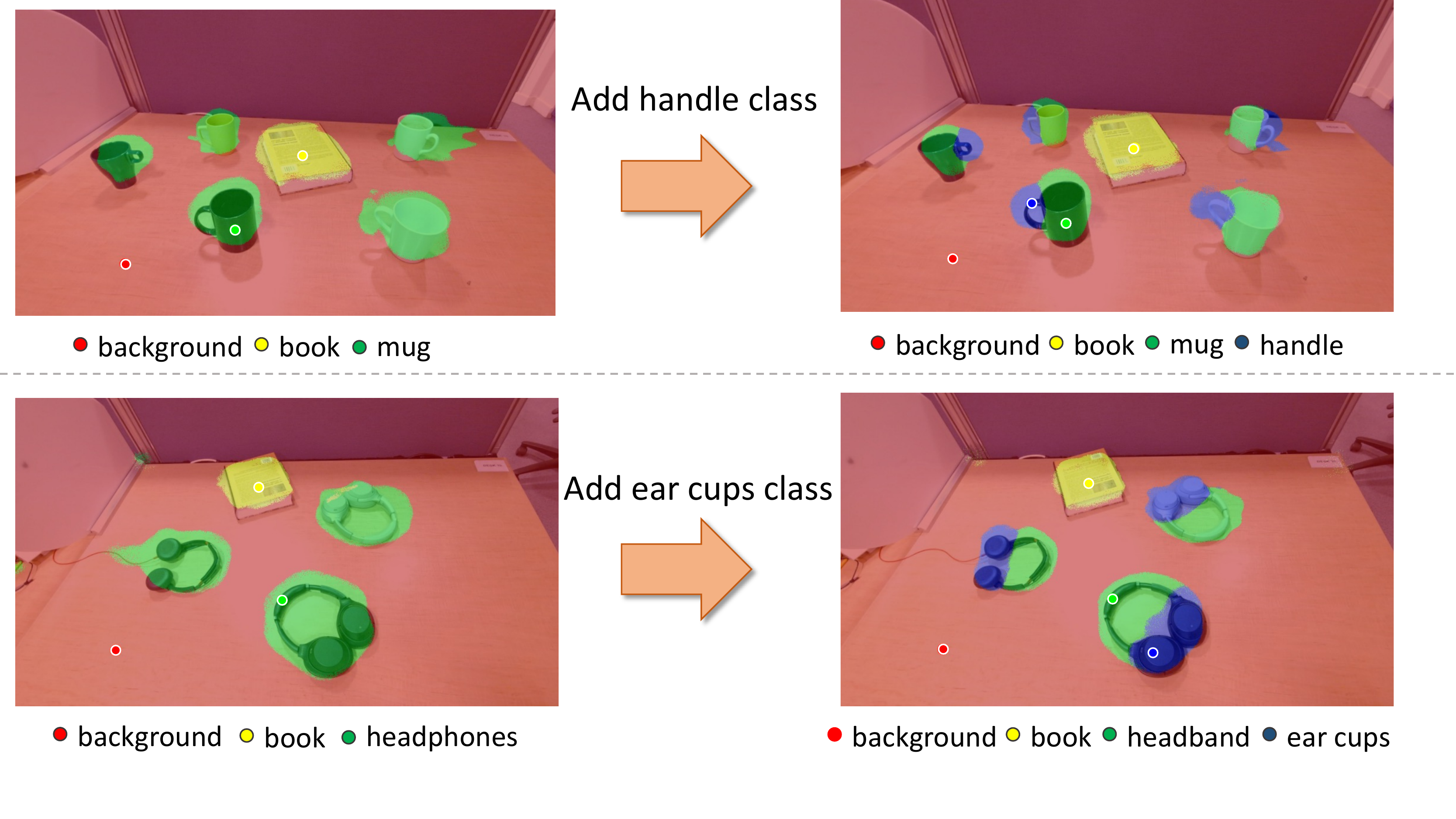}
    \caption{\textbf{Specialization.} In these experiments we show that if one click is added to an object to indicate a new sub-part class, all instances are correctly divided into the same part classes.}
    \label{fig:specialization}
\end{figure}

First, as in the previous experiment we assign one label to the target object instance, one to the background, and one to the unrelated object. In this initial step we observe  similar behaviour to before: our system groups the target objects together, while separating out the background and the additional book class. 

In the second stage a new label is introduced, denoting a subclass of the initial class: handle and ear cover for the base mug and headphones classes respectively. The results are illustrated in \Cref{fig:specialization}. We observe that the system successfully dissects one object class into two and propagates this dissection to all object instances of the same class, indicating useful part or affordance representation capabilities.

\subsection{Exploration}
\label{sec:exploration}
\begin{figure*}[h!]
  \includegraphics[width=\textwidth]{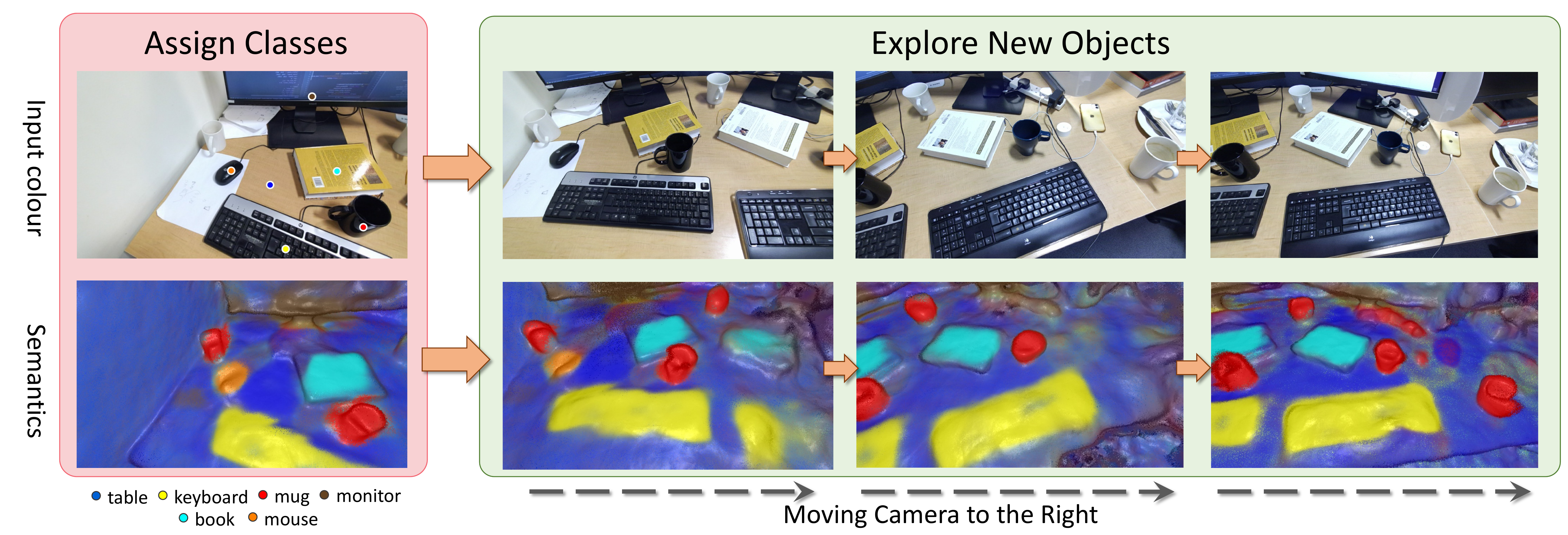}
  \caption{\textbf{Exploration.} Objects are annotated with one click each in the first frame; dense segmentation then correctly propagates to new instances as the camera explores and the network representing reconstruction and features trains continuously. 
  \label{fig:exploration}
  }
\end{figure*}

We also test in an exploratory setting where only part of the whole scene is visible and reconstructed at the label assignment stage:
see 
\Cref{fig:exploration}. First, labels are assigned in the first view to one object per class as in the experiments of \Cref{sec:coverage}. Then, the camera moves to view new parts of the scene with other objects. The scene network adds new keyframes automatically and continues to train its geometry/feature representation in real-time.
In \Cref{fig:exploration} we can  observe how the correct semantic labels ``emerge'' automatically on newly observed objects as they are reconstructed. 
Our method strongly propagates labels to object instances unseen at the label assignment stage: mugs, books, keyboard, monitor bases, and unobserved table regions.

\subsection{Feature Extractor}
\label{sec:cnn_vs_vit}

\begin{figure}[h!]
    \centering
    \includegraphics[width=\linewidth, 
                     trim={0 11.0cm 8cm 0},clip]{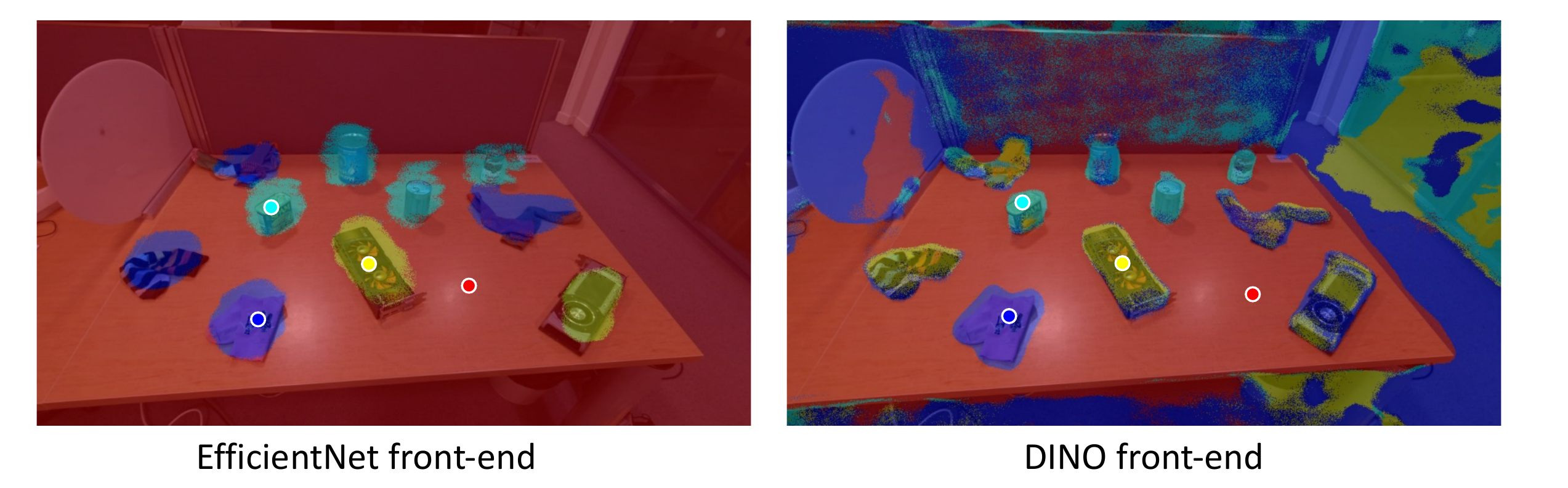}
    \caption{\textbf{Feature Front-End Choice}. While the DINO-based feature front-end yields more geometrically-accurate masks, they have weaker semantic entanglement and some objects are ambiguously labelled.}
    \label{fig:cnn_vs_vit}
\end{figure}

Despite the emerging trend~\cite{melaskyriazi:etal:CVPR2022, amir:etal:ARXIV2021} of leveraging DINO for  unsupervised semantic segmentation, we have observed both qualitatively and quantitatively that a supervised CNN yields stronger results in our setting.

Our qualitative (see \Cref{fig:cnn_vs_vit}) and quantitative evaluations indicate an interesting trade off between these two front-ends: a Transformer-based feature extractor provides cleaner semantic boundaries for objects, whereas a CNN-based facilitates stronger semantic entanglement for complex objects at the expense of geometric accuracy.  We would expect that if the SLAM reconstruction of our system was more precise, the CNN front-end would also be able to get similarly accurate semantic boundaries.

\subsection{Quantitative Evaluation}
\label{sec:quantitative}
\begin{figure}[h!]
    \centering
    \includegraphics[width=\linewidth, 
                     ]{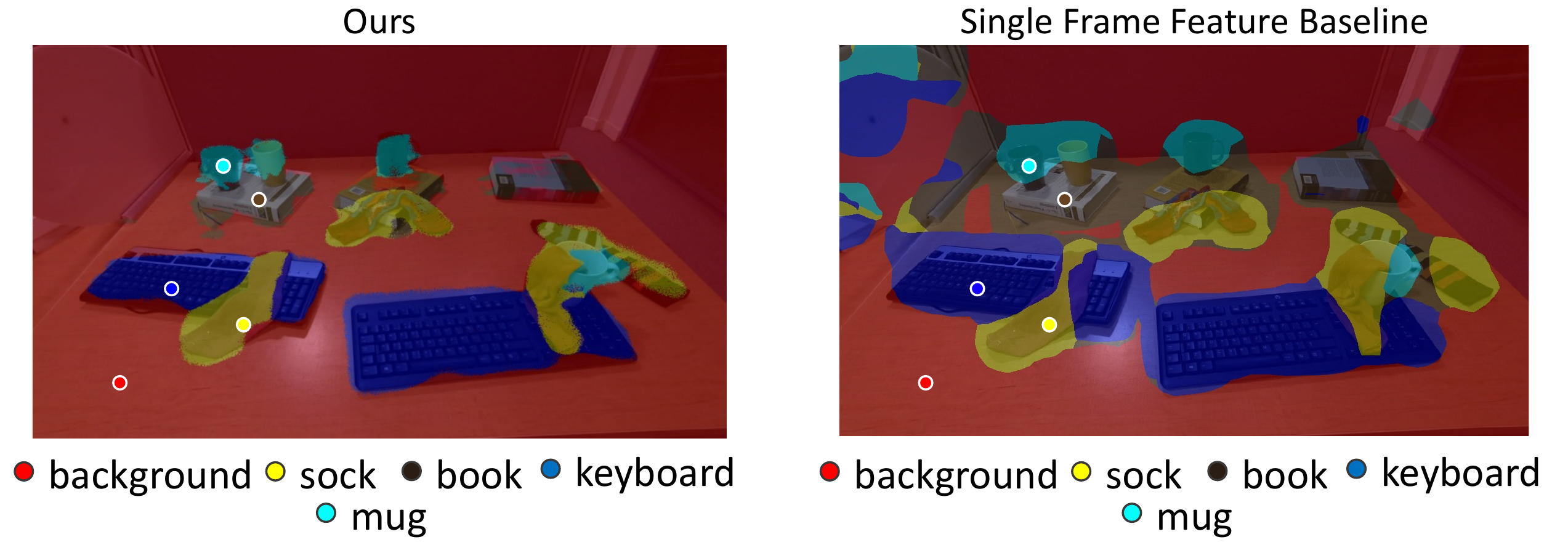}
    \caption{\textbf{Qualitative comparison with baseline.} We also evaluate qualitatively how our model performs compared to the pure feature-based baseline discussed in~\Cref{sec:quantitative}. Note that our method yields sharper boundaries compared to the baseline.}
    \label{fig:clutter_baseline}
\end{figure}

To evaluate the performance of our method quantitatively, we adopt a standard Mean Intersection over Union (mIoU) to measure the semantic segmentation quality of our method. We annotate some image regions as an ignore class (desk separators, single instance clutter, and other regions outside the tabletop) to only measure how well does our method handles semantic object grouping.

Since our method is complementary to and disentangled from front-end feature quality, we compare it with an unfused, purely 2D feature-based, method. 
In our protocol the target classes are defined via a single click, i.e a labelled pixel. Therefore, we devise a baseline which measures feature similarity between the features  at a target pixel and features associated with the labelled pixels. 

Recent work~\cite{melaskyriazi:etal:CVPR2022} has demonstrated the strong performance of clustering DINO features for semantic segmentation. Inspired by this approach we design our baselines using feature-metric clustering. Let $\bf{I}_0$ be the starting image from a test sequence with a set of user-defined clicks $\{(\zeta_i, c_i)\}$ on it, where $\zeta_i \in [0, 1]^2$ is the spatial position of a pixel click and $c_i$ is its class label. We query $\mathcal{F}(\mathbf{I_0})$ using bilinear interpolation at spatial positions $\zeta_i$ to obtain anchor features $f_i = \mathcal{F}(\mathbf{I_0})[\zeta_i]$. To classify pixels in a target image $\mathbf{I_j}$, we pass it through the feature extractor $\mathcal{F}(\mathbf{I_j})$ and assign a class label in a one-nearest-neighbour classification manner, for each pixel based on the feature-metric distance to the anchor set $\{f_i\}_i$. Cosine similarity performed best in our case, which is consistent with the literature.

Furthermore we show quantitatively the benefits of our introduction of learned priors. We therefore compare the performance with that of iLabel using the same clicks. Unlike our system, iLabel does not use any prior learned information and relies purely on colour and geometry self-similarities in the underlying neural scene model.

Our method improves over the pure feature-based counterpart by a large margin for both EfficentNet and DINO front-ends. We argue that this is due its noise-filtering and spatial upsampling properties. Interestingly, the DINO front-end outperforms EfficentNet on the desk sequence, with more common (mugs, keyboards, books) classes. This due to the fact that the DINO-based front-end provides weaker semantic entanglement, yet produces finer semantic masks, as has already been discused in~\Cref{sec:cnn_vs_vit}. Meanwhile EfficentNet thrives in settings with less common objects, such as GPUs, socks, plants, etc.

\begin{table}[h]
\setlength{\tabcolsep}{2.5pt}
\begin{tabular}{ |c | c c c c c| c|  }
 \hline
 Method &  Plants & Desk & Sponge & Trainers & GPUs & Mean\\
 \hline
 \textbf{Fused EffcientNet (ours)} &   \textbf{65.1} & 57.8 & \textbf{59.9} & \textbf{65.4} & \textbf{59.4} & \textbf{61.5} \\
 \textbf{Fused DINO (ours)}  & 46.4 & \textbf{63.3} & 56.6 & 42.1 & 52.5 & 52.2 \\
  \hline
 EffiecentNet baseline   & 41.3 & 50.0 & 38.1 & 47.5 & 51.6 & 43.7 \\
  DINO baseline & 40.4 & 59.1 & 55.4 & 45.4 & 39.4 & 45.9 \\
  \hline
 iLabel  & 46.4 & 30.6 & 32.9 & 37.9 & 27.1 & 35.0 \\
  \hline
\end{tabular}
\caption{\textbf{mIOU scores.} Quantitative evaluation of our Feature-Realistic Fusion system performance on our tabeltop sequences. Feature-Realistic Fusion demonstrates consistent improvement over pure feature-based baselines for both vision front-ends as well as over the colour-based iLabel system.}
  \label{tab:quantitative}
\end{table}

\subsection{Limitations}

\begin{figure}[h]
    \centering    \includegraphics[width=\linewidth]{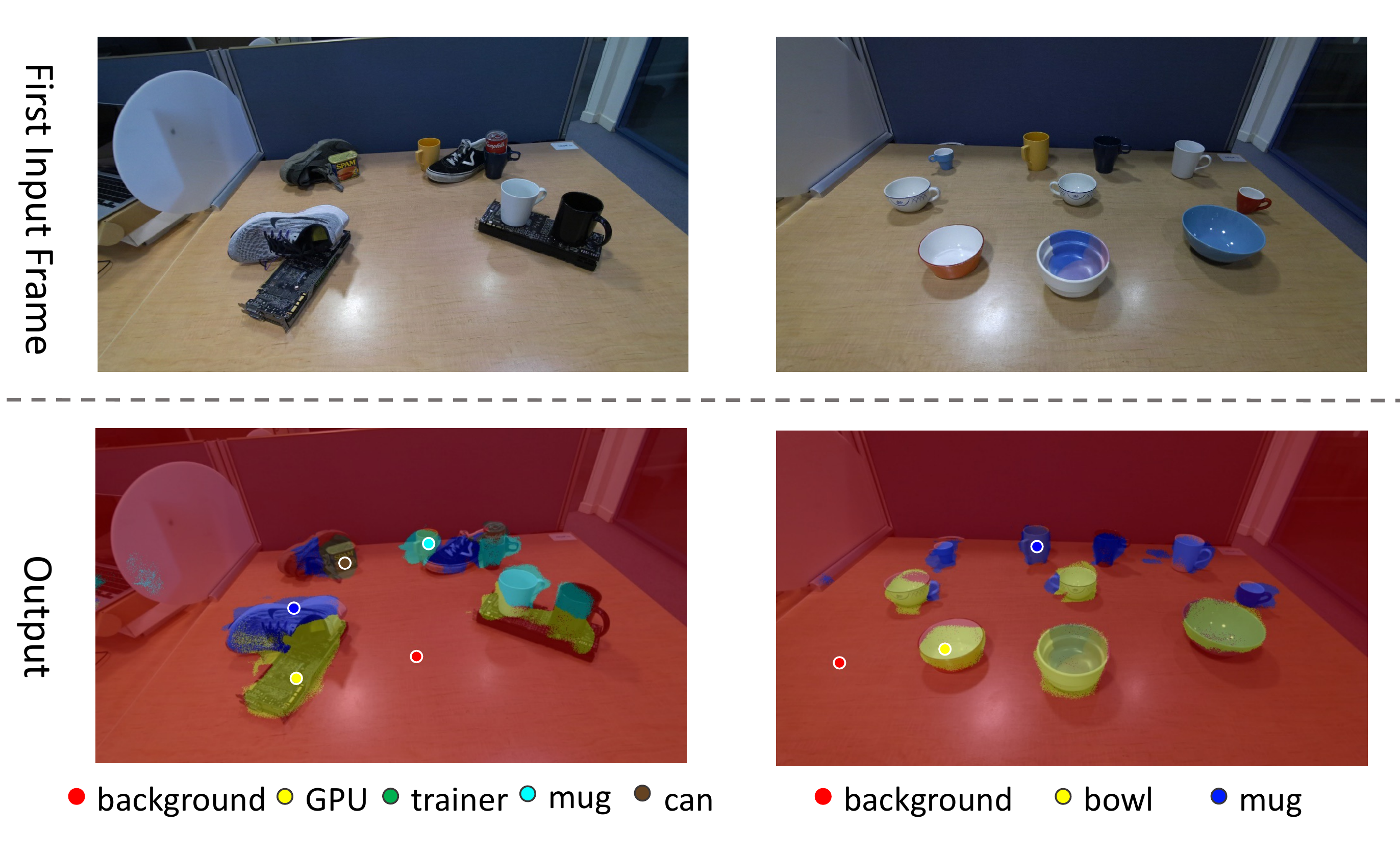}
    \caption{\textbf{Limitations.} On the left, segmentation accuracy is degraded in a highly complex scene with unusual compound objects such as GPUs and shoes; on the right, extremely similar classes such as mugs and bowls can be confused.}
    \label{fig:limitations}
\end{figure}

\label{sec:limitations}

We provide a qualitative example in \Cref{fig:limitations} (left) of where our method's pixel accuracy in a cluttered scene severely degrades (yet the label assignment remains valid) due to the presence of rare classes. Meanwhile in \Cref{fig:limitations} (right) our method struggles to differentiate mugs with a wide body from bowls due to their natural semantic connection. 

%% file: conclusion.tex
\section{Conclusion}

We have shown that real-time fusion of general high-dimensional features can be efficiently and simply achieved within a neural field SLAM system, and that this enables scenes to be densely semantically segmented with only a tiny amount of run-time, open-set annotation.
This approach is particularly promising for robotics in complex and unusual domains where pre-trained semantic segmentation networks currently perform poorly, and we plan to soon run it at larger scale and with more complex scenes.